\documentclass[sigconf,natbib=false]{acmart}
\AtBeginDocument{%
  }

\copyrightyear{2026}
\acmYear{2026}
\setcopyright{cc}
\setcctype{by}
\acmConference[SAC '26]{The 41st ACM/SIGAPP Symposium on Applied Computing}{March 23--27, 2026}{Thessaloniki, Greece}
\acmBooktitle{The 41st ACM/SIGAPP Symposium on Applied Computing (SAC '26), March 23--27, 2026, Thessaloniki, Greece}
\acmPrice{}
\acmDOI{10.1145/3748522.3779909}
\acmISBN{979-8-4007-2294-3/2026/03}

\RequirePackage[
  datamodel=acmdatamodel,
  style=acmnumeric,
  sorting=none
]{biblatex}

\begin{document}

\title[\textsc{EntroLnn}: Entropy-Guided Liquid Neural Networks]{\textsc{EntroLnn}: Entropy-Guided Liquid Neural Networks for Operando Refinement of Battery Capacity Fade Trajectories}


\author{Wei	Li}
\affiliation{%
  \institution{Singapore Institute of Technology}
  \country{Singapore}}
\email{wei.li@singaporetech.edu.sg}

\author{Wei Zhang}
\authornote{Corresponding author.}
\affiliation{%
  \institution{Singapore Institute of Technology}
  \country{Singapore}}
\email{wei.zhang@singaporetech.edu.sg}

\author{Qingyu Yan}
\affiliation{%
  \institution{Nanyang Technological University}
  \country{Singapore}}
\email{alexyan@ntu.edu.sg}

\renewcommand{\shortauthors}{W. Li et al}

\begin{abstract}
Battery capacity degradation prediction has long been a central topic in battery health analytics, and most studies focus on state of health (SoH) estimation and end of life (EoL) prediction. This study extends the scope to online refinement of the entire capacity fade trajectory (CFT) through \textsc{EntroLnn}, a framework based on \underline{entro}py-guided transformable liquid neural networks (\underline{LNN}s). \textsc{EntroLnn} treats CFT refinement as an integrated process rather than two independent tasks for pointwise SoH and EoL. We introduce entropy-based features derived from online temperature fields, applied for the first time in battery analytics, and combine them with customized LNNs that model temporal battery dynamics effectively. The framework enhances both static and dynamic adaptability of LNNs and achieves robust and generalizable CFT refinement across different batteries and operating conditions. The approach provides a high fidelity battery health model with lightweight computation, achieving mean absolute errors of only 0.004577 for CFT and 18 cycles for EoL prediction. This work establishes a foundation for entropy-informed learning in battery analytics and enables self-adaptive, lightweight, and interpretable battery health prediction in practical battery management systems.
\end{abstract}




\maketitle

\section{Introduction}
Capacity degradation is an inevitable process in lithium-ion batteries (LIBs) and has long been a central topic in their application and development \cite{Sheng_Zhang}. Recent studies have shown advances in battery degradation prediction in terms of the estimations of state of health (SoH), remaining useful life (RUL), and end of life (EoL) \cite{Xiankui_Wu, Ganglin_Cao}. In general, the methodological focus of battery health modelling has gradually shifted from traditional data-driven methods, such as Gaussian process regression, to more advanced machine-learning (ML) algorithms that can model nonlinear and high-dimensional behaviors \cite{Meng_Xu, Demirci}. More recently, purely data-driven models have evolved into physics-informed ML frameworks, which integrate physical knowledge and ML to enhance modelling accuracy, robustness, and interpretability \cite{Guoqing_Sun}. These advancements show a transition from empirical parameter fitting to ML models capable of capturing complex electrochemical degradation patterns. However, significant challenges remain in developing generalizable, interpretable, and sensor-efficient models for battery degradation. In practice, batteries operate under diverse operating conditions that may differ from those in controlled laboratory environments. Most existing models rely on complete measurements of current, voltage, and cycle data, and often assume homogeneous operating conditions; consequently, they tend to lose accuracy and reliability in real systems. To support practical battery analytics, degradation models must be both data-efficient and physically meaningful, with operando adaptability to track degradation under real-world conditions \cite{Tingkai_Li}.

Several recent approaches have considered the above challenges by improving data efficiency and incorporating physical constraints into learning models, and have achieved progresses in SoH and EoL prediction accuracy \cite{Yifan_Zheng}. In \cite{Yin-Yi_Soo}, one-dimensional convolutional neural networks (1D-CNNs) and Transformer models are applied to model temporal dependencies within batteries and estimate SoH from partial cycle data. Bidirectional long short-term memory (Bi-LSTM) networks are employed in \cite{Tingting_Tao} to reconstruct capacity-fade trajectories (CFTs) to provide a more continuous representation of degradation compared with single-point estimation. In parallel, liquid neural networks (LNNs) \cite{Ramin_Hasani_NML} have emerged as a promising ML algorithm for modeling dynamic processes through continuous-time state evolution. While good performance has been reported for these models, their scalability remains limited with insufficient physical grounding, and the modeled degradation patterns are specialized to certain datasets without sufficient generalizability \cite{Yaodi_Huang}. To enhance generalization, recent models have been developed to integrate electrochemical insights with ML \cite{Aihua_Tang, Chuang_Chen}. These models combine the flexibility of data-driven learning with physical modeling to achieve CFT refinement without being dominated by specific data distributions. However, these models still rely on large datasets that span diverse operating conditions, since their embedded electrochemical insights are themselves parameterized from data. As such, their learned physical constraints often fail to generalize under unseen conditions, and the resulting representations remain largely empirical. This reflects a deep conceptual limitation, that is the absence of universal and physically meaningful features that can quantify degradation consistently across diverse batteries and operating conditions.

\begin{figure*}[!t]
\centering
\includegraphics[width=\textwidth]{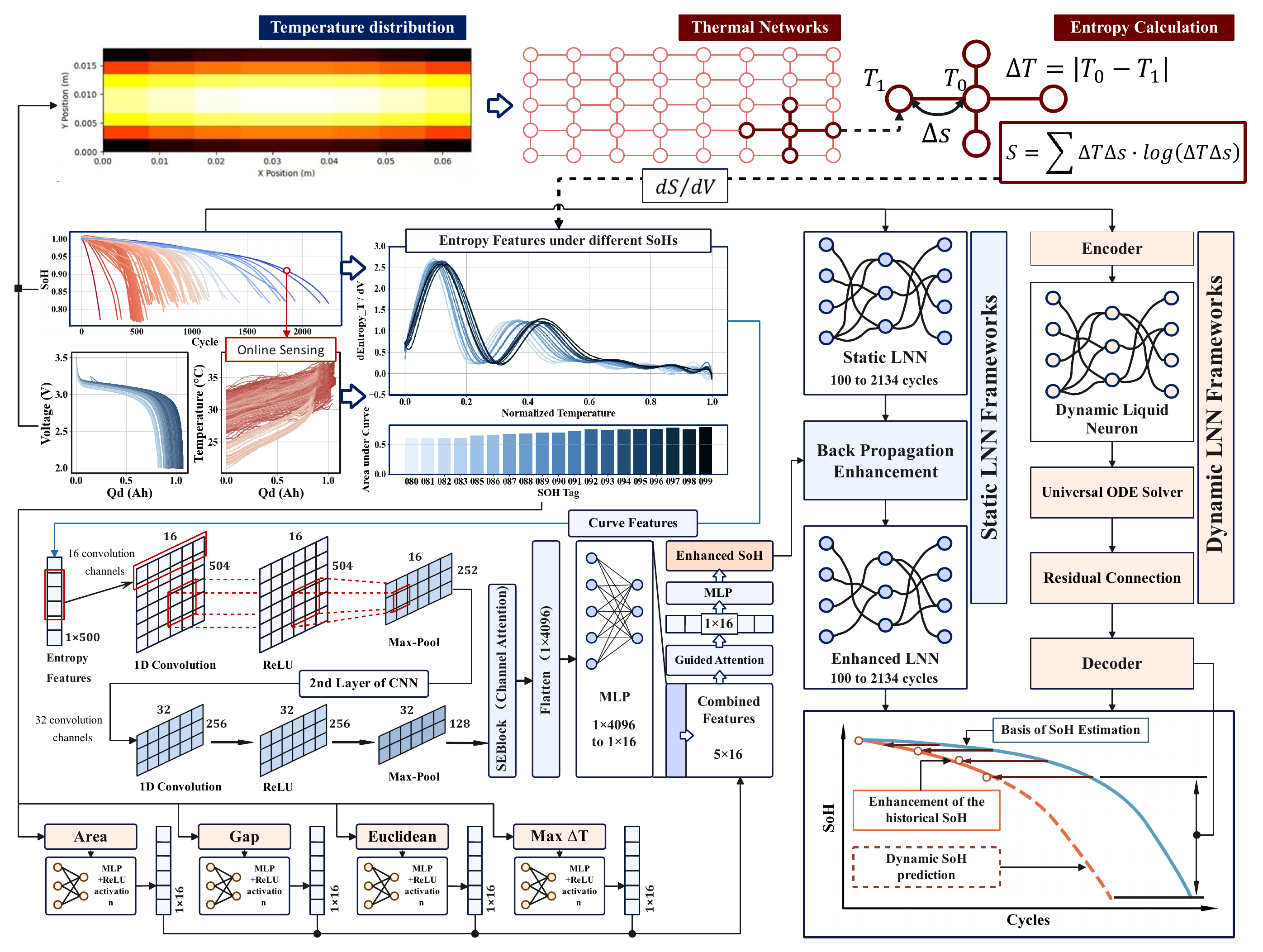}
\caption{The holistic framework of \textsc{EntroLnn}. The model integrates multiple features of LIBs, including temperature curves, entropy features derived from temperature profiles, and additional attributes extracted from entropy curves. \textsc{EntroLnn} adapts to real-time CFT refinement across different batteries and maintains high generalizability and stable performance.
}
\label{fig:EntroTL}
\end{figure*}

To address the absence, we introduce \emph{entropy} as a set of statistical features of internal degradation of batteries. Our proposed entropy is inspired by information entropy, which quantifies the uncertainty within a distribution. We extend this principle to describe the spatial and temporal irregularity of the battery temperature field during degradation, where the irregularity tends to increase as degradation progresses. In this way, entropy provides a physically inspired and thermally interpretable descriptor to reflect the growing temperature heterogeneity during battery aging. Unlike previous models that use data-parameterized electrochemical equations, our proposed entropy is derived directly from measured temperature. The entropy is insensitive to short-term fluctuations like noises and highlights consistent thermal patterns associated with degradation. As such, entropy in this study serves as an interpretable and operando-computable indicator that offers a unified representation connecting the evolving temperature field and degradation for real-time CFT refinement.

Based on the proposed entropy formulation, we develop \textsc{EntroLnn}, the \underline{entro}py-guided transformable \underline{LNN}s designed to refine battery CFTs in real-time. The overall workflow consists of two main components, including constructing entropy-based degradation indicators and training transformable LNNs. Ambient temperature and current measurements are first used to reconstruct the temperature field of a battery. The field is then converted into entropy features that quantify the dispersion of heat generation across space and time. These features, together with voltage and temperature data, form an entropy-based SoH indicator that captures the joint evolution of thermal variability and battery degradation. The indicator establishes an interpretable link between entropy and measurable battery capacity loss and serves as the physical input to our LNN models. In the first stage, a static LNN is trained using the data of a single long-lifespan reference battery, which has stable early-cycle degradation dynamics and can be the foundation for CFT representation. The static LNN cannot perfectly generalize to entire CFTs. So, we train a dynamic LNN using capacity-fade data from several batteries with different degradation profiles to learn the temporal evolution of degradation until the EoL of each battery. During \textsc{EntroLnn} deployment, the two LNNs operate jointly. Early-cycle data, e.g., the first 100 cycles of SoH and entropy-based features, are used to fine-tune the static LNN for generating CFT, which is primarily optimized for the early degradation stage. When the SoH decreases beyond, for example, 98\%, the dynamic LNN adapts to operating conditions and extends the refined trajectory until EoL.

We have conducted extensive experimental studies using a public dataset, and the proposed \textsc{EntroLnn} demonstrates strong performance in both accuracy and adaptability across different batteries. With only entropy-based features and limited early-cycle data, \textsc{EntroLnn} accurately reconstructs and refines the CFTs of unseen batteries throughout their lifespans. The incorporation of entropy features significantly enhances modeling accuracy and robustness by filtering short-term fluctuations and noises to generate physically consistent degradation trajectories. \textsc{EntroLnn} achieves mean absolute errors (MAEs) below 0.003 while maintaining high computational efficiency and strong generalization across different operating conditions. Beyond quantitative improvements, \textsc{EntroLnn} establishes a new paradigm for accurate and robust battery health modeling. It bridges the gap between data-driven learning and physically inspired representation, and enables real-time and generalizable CFT refinement from minimal data. Furthermore, the entropy concept provides a transferable foundation for broader energy-related applications, and advances physics-aware machine intelligence and sustainable energy technologies.

The remainder of this paper is structured as follows. Section \ref{sec:method} introduces the methodology of \textsc{EntroLnn}. Sections \ref{sec:data} and \ref{sec:experiment} present data and the experimental study and results, respectively. Finally, section \ref{sec:conclusion} summarizes this paper.

\section{\textsc{EntroLnn}: Methodology}
\label{sec:method}
We present the \textsc{EntroLnn} methodology, as shown in Figure \ref{fig:EntroTL}.

\subsection{Entropy-Guided Online Feature Extraction}
The temperature field serves as an innovative source of physical information for CFT refinement. However, directly incorporating the raw field information into ML models introduces excessive dimensionality, which presents significant challenges for learning. Moreover, spatial correlations among temperature points are not explicitly captured in the raw field. To address these limitations, we develop entropy-based features that summarize the overall spatial–thermal characteristics of the temperature field, and provides a compact and physically interpretable descriptor of thermal non-uniformity associated with battery degradation. The entropy $S$ is given by,
\begin{equation}
\label{eq:Entropy_define}
S = \sum \Delta T \Delta s \cdot \log(\Delta T \Delta s),
\end{equation}
where the entropy encapsulates the joint information of temperature $T$ and spatial intervals $\Delta s$, to effectively quantify the degree of thermal non-uniformity in the battery that evolves with degradation. $\Delta T$ denotes the absolute temperature difference, which ensures that the calculated values remain non-negative. The underlying physics is illustrated in Figure \ref{fig:EntroTL}. Notably, the entropy $S$ follows the mathematical form of information entropy but serves here to indicate battery-specific characteristics. Unlike Shannon entropy, which measures statistical uncertainty, this entropy quantifies each battery’s thermal dispersion. Hence, normalization is not adopted as it would remove physically meaningful magnitude differences across conditions. According to the simulation results, the accuracy of temperature calculations will not be significantly improved when the number of nodes in the temperature field exceeds 8$\times$8, which is selected to calculate the entropy in this study. Subsequently, an operando physics-informed online adaptation is applied to the heat generation model to mitigate errors associated with capacity degradation. The underlying mechanism of this method is derived from the basic thermal model \cite{Alkhedher} as,
\begin{equation}
\label{eq:basic thermal}
f\big(T(t); k\big) \triangleq C_p m \frac{\mathrm{d}T}{\mathrm{d}t} - kQ(t) = 0,
\end{equation}
which is a functional representation of the general temperature curve from a lumped thermal model in which convection is neglected for simplicity. Here, $C_p$ denotes the specific heat capacity, $m$ is the mass of the battery, and $Q(t)$ represents the heat generation. The coefficient $k$ is introduced to calibrate the thermal generation term. Based on the physical dynamics in Eq. (\ref{eq:basic thermal}), the coefficient $k$ directly influences the evolution of temperature curves. The operando adaptation of $k$ aims to minimize the discrepancy between the predicted and actual temperature curves through the following equations,
\begin{equation}
\label{eq:k_calibration}
\mathcal{L}_T = \left( T-T^* \right)^2,\quad k \triangleq {k} -  \eta\frac{\mathrm{d} \mathcal{L}_T}{\mathrm{d}{k}},
\end{equation}
where $\mathcal{L}_T$ denotes the distance between the practical and reference temperatures $T$ and $T^*$, respectively, and $k$ is derived based on the distance with learning rate $\eta$. Based on the chain rule of differentiation, we have,
\begin{equation}
\label{eq:loss_chain}
\frac{\mathrm{d} \mathcal{L}_T}{\mathrm{d}{k}} = \frac{\mathrm{d} \mathcal{L}_T}{\mathrm{d}T}\frac{\mathrm{d}T}{\mathrm{d}{k}},
\end{equation}
where ${\mathrm{d}\mathcal{L}_T}/{\mathrm{d}T}$ equals to $2\delta_T$ and $\delta_T$ represents the temperature difference $T-T^*$. Based on Eq. (\ref{eq:basic thermal}), ${\mathrm{d}T}/{\mathrm{d}{k}}$ can be given as,
\begin{equation}
\label{eq:dTdk}
\frac{\mathrm{d}T}{\mathrm{d}{k}} = \int_{0}^{t}  \frac{Q(t)}{C_p m} \mathrm{d}t.
\end{equation}
It follows that $\frac{\mathrm{d}T}{\mathrm{d}{k}}$ is independent of $k$ but depends on the temperature evolution over the observation window, and the operando adaptation of ${k}$ is reformulated as follows:
\begin{equation}
\label{eq:k_adaption}
{k}\triangleq {k} -  \eta^*\delta_T,
\end{equation}
where $\eta^*$ is the generalized learning rate corresponding to the temperature difference. Finally, our simulation study demonstrates that the derivative of entropy with respect to voltage $\mathrm{d}S/\mathrm{d}V$ exhibits a strong correlation with SoH as shown in Figure \ref{fig:EntroTL}. Therefore, the derivative $\mathrm{d}S/\mathrm{d}V$ is adopted as the terminal temperature-based entropy feature for CFT refinement. 

\subsection{Entropy-Guided SoH Modelling}
As illustrated in Figure \ref{fig:EntroTL}, the derivative of entropy with respect to voltage ($\mathrm{d}S/\mathrm{d}V$) serves as the primary feature for modelling SoH. To extract representative patterns from this curve, a lightweight two-layer 1D-CNN \cite{Alharbi} is employed. The ML model encodes the local structure and variation of the $\mathrm{d}S/\mathrm{d}V$ profile, which reflects the coupled thermal–degradation behavior of the battery. The overall feature-extraction process can be formulated as,
\begin{equation}
\label{eq:CNN}
\left\{
\begin{aligned}
\mathbf{X}^{(1)} &= \texttt{ReLU}\big(  \mathbf{W}^{(1)}_{C}\ast\mathbf{x}^{(0)} + \mathbf{b}^{(1)}_{C}\big), \\
\mathbf{X}^{(2)} &= \texttt{MaxPool}\big(\mathbf{X}^{(1)}\big), \\
\mathbf{X}^{(3)} &= \texttt{ReLU}\big(  \mathbf{W}^{(2)}_{C}\ast\mathbf{X}^{(2)} + \mathbf{b}^{(2)}_{C}\big), \\
\mathbf{X}^{(4)} &= \texttt{MaxPool}\big(\mathbf{X}^{(3)}\big), 
\end{aligned}
\right.,
\end{equation}
where $\mathbf{x}^{(0)}$ denotes the input vector corresponding to the sequential $\mathrm{d}S/\mathrm{d}V$ curve, uniformly resampled to 500 points along the temperature axis where the temperatures are normalized. $\mathbf{W}^{(1)}_{C}$ and $\mathbf{b}^{(1)}_{C}$ are the weights and biases of the first layer, and $\mathbf{W}^{(2)}_{C}$ and $\mathbf{b}^{(2)}_{C}$ represent those in the second layer. The channel numbers of each convolutional layer are 16 and 32, respectively. The kernel sizes of each convolutional layer and max-pooling layer are 5 and 2, respectively, in this study.

While the $\mathrm{d}S/\mathrm{d}V$ feature provides the dominant entropy-based representation of the battery’s coupled thermal–degradation behavior, additional physics-informed indicators are incorporated to enhance robustness and interpretability. These include the area under the $\mathrm{d}S/\mathrm{d}V$ curve, the gap between its two dominant peaks, and the Euclidean distance between each $\mathrm{d}S/\mathrm{d}V$ curve and the reference profile, all derived directly from the $\mathrm{d}S/\mathrm{d}V$ curve. To further improve cross-validation performance in SoH modelling, the maximum temperature increment $\Delta T_{\max}$ is also incorporated as an external thermal indicator. Together, these complementary features capture distinct thermal and electrochemical degradation characteristics and form a compact multi-physics feature set for entropy-guided SoH modelling.  
All extracted features are first projected into a 16-dimensional latent space and fused through an attention mechanism that adaptively weighs their relative importance. The fused representation is subsequently passed to a two-layer multilayer perceptron (MLP) to generate the final SoH estimate, expressed as,
\begin{equation}
\label{eq:Feature fusion}
\left\{
\begin{aligned}
& \mathbf{x}_{i} = \texttt{ReLU}(\mathbf{W}_{i} \cdot S_i + \mathbf{b}_{i}), \quad i = 1,\dots,5, \\
& \mathbf{z} = \texttt{concat}(\mathbf{x}_1,\mathbf{x}_2,\ldots,\mathbf{x}_5), \\
& \mathbf{w} = \texttt{softmax}(\mathbf{W}_{\text{att}}\mathbf{z}+\mathbf{b}_{\text{att}}), \\
& \mathbf{x}_{\text{fused}} = \sum\nolimits_{i=1}^{5} w_i \mathbf{x}_i, \\
& \hat{y}_{\text{SoH}} = \mathbf{W}'_2 \!\cdot\! \texttt{ReLU}(\mathbf{W}'_1 \mathbf{x}_{\text{fused}} + \mathbf{b}'_1) + b'_2,
\end{aligned}
\right.,
\end{equation}
where $S_i$ represents the $i$-th physics-informed feature and $\mathbf{x}_i \in \mathbb{R}^{16}$ is its latent representation. $\mathbf{W}_i$ and $\mathbf{b}_i$ denote the trainable weight matrix and bias vector for the first projection layer. $\mathbf{w} = [w_1, \ldots, w_5]$ is the attention weight vector obtained via the \texttt{softmax} layer. $\mathbf{x}_{\text{fused}}$ is the aggregated embedding of entropy features. Finally, the estimated SoH, $\hat{y}$, is produced by the two-layer MLP parameterized by $\mathbf{W}'_1$, $\mathbf{b}'_1$, $\mathbf{W}'_2$, and $b'_2$.

\subsection{LNNs for Operando CFT Refinement}
To address the diversity of degradation behaviors across different batteries, an operando-adaptive framework is developed that combines static and dynamic LNNs. The static LNN captures short-term degradation dynamics and enables online refinement under new operating conditions, while the dynamic LNN models long-term evolution and captures nonstationary degradation patterns with higher adaptability. Together, they form a unified framework for continuous refinement of CFT in real-time. The static LNN is first trained on a reference battery with long lifespan using only the first 100 cycles of the battery as input to reconstruct the complete CFT. During deployment, this model serves as the foundation for operando adaptation to other batteries with limited early-cycle data. The encoding process initializes the hidden state of the LNN as,
\begin{equation}
\label{eq:encoder_LNN}
\mathbf{h}_0 = \mathbf{W}_{\text{enc}} \cdot \mathbf{x}_{\text{in}} + \mathbf{b}_{\text{enc}}, 
\quad \mathbf{h}_0 \in \mathbb{R}^{64},
\end{equation}
where $\mathbf{x}_{\text{in}}$ denotes the SoH sequence from the first 100 cycles, 
and $\mathbf{W}_{\text{enc}}$ and $\mathbf{b}_{\text{enc}}$ are the encoder weights and bias.
$\mathbf{h}_0$ serves as the initial hidden state of the LNN. The hidden-state dynamics of the static LNN follow the standard continuous-time ordinary differential equation (ODE) form \cite{Ramin_Hasani} as,
\begin{equation}
\label{eq:hidden states_LNN}
\frac{\mathrm{d}\mathbf{h}}{\mathrm{d}t} = 
-\boldsymbol{\alpha} \odot \mathbf{h} + \tanh\left(\mathbf{W}_{h}\mathbf{h} + \bar{\mathbf{u}}\right),
\end{equation}
where $\mathbf{h}$ is the hidden-state vector initialized as $\mathbf{h}_0$, $\boldsymbol{\alpha}$ is the trainable decay coefficient, and $\mathbf{W}_{h}$ denotes the recurrent-weight matrix. In conventional LNN formulations, $\mathbf{u}(t)$ denotes the time-varying external input that influences the hidden-state dynamics. In this study, we let $\mathbf{u}(t)$ represent the operational conditions such as temperature or current. For model compactness, we introduce $\bar{\mathbf{u}}$ as the averaged early-cycle input as a static proxy for $\mathbf{u}(t)$ as,
\begin{equation}
\label{eq:constant_LNN}
\bar{\mathbf{u}} = \frac{1}{100}\sum\nolimits_{t=1}^{100}\mathbf{x}_{\text{in}}^{(t)}.
\end{equation}
The hidden state $\mathbf{h}$ evolves with respect to time and is integrated numerically using a Runge–Kutta ODE solver with adaptive step size as,
\begin{equation}
\label{eq:solution of ODE}
\mathbf{H}(t) = 
\texttt{odeint}\!\left(\tfrac{\mathrm{d}\mathbf{h}}{\mathrm{d}t},\;
\mathbf{h}_0,\; t\right).
\end{equation}
Finally, the output layer maps the hidden states to the corresponding SoH estimates as,
\begin{equation}
\label{eq:output_LNN}
\hat{\mathbf{y}}_{\text{SoH}} = \mathbf{W}_{\text{out}}\mathbf{H}(t_T) + \mathbf{b}_{\text{out}},
\end{equation}
where $\mathbf{W}_{\text{out}}$ and $\mathbf{b}_{\text{out}}$ are the output weights and bias, and $t_T$ denotes the final cycle identifier corresponding to the battery's EoL. The static LNN is trained based on a reference battery and we aim to improve its generalizability by updating its parameters during deployment through operando adaptation. Let $\boldsymbol{\theta}$ be the training parameters of the static LNN. The online parameter refinement follows a gradient-based update rule as,
\begin{equation}
\label{eq:operando refinement_LNN}
\boldsymbol{\theta} \leftarrow 
\boldsymbol{\theta} - \eta \nabla_{\boldsymbol{\theta}}\mathcal{L}_{\text{total}},
\end{equation}
where $\eta$ is the learning rate and $\nabla_{\boldsymbol{\theta}}\mathcal{L}_{\text{total}}$ 
represents the gradient of the total loss with respect to $\boldsymbol{\theta}$. The loss $\mathcal{L}_{\text{total}}$ combines a value term and a slope-matching term to jointly constrain pointwise accuracy and trajectory smoothness, and is given by,
\begin{equation}
\label{eq:total loss}
\mathcal{L}_{\text{total}}
= \mathcal{L}_{\text{val}} + 
\gamma \mathcal{L}_{\text{slo}},
\end{equation}
where $\gamma$ controls the relative importance of the slope term and the two loss components are defined as,
\begin{equation}
\label{eq:value_slope_loss}
\mathcal{L}_{\text{val}} = \frac{1}{T}\sum\nolimits_{t=1}^{T} (\hat{y}_t - y_t)^2, ~ \mathcal{L}_{\text{slo}} = \frac{1}{T-1}\sum\nolimits_{t=1}^{T-1} \big(\nabla \hat{y}_t - \nabla y_t\big)^2,
\end{equation}
where $\hat{y}_t$ and $y_t$ are the predicted and measured SoH values at cycle $t$, 
and $\nabla \hat{y}_t$ denotes the discrete derivative of the predicted SoH curve.
This formulation ensures that the adapted model not only fits the measured values 
but also preserves consistency of degradation trends.

The static LNN is designed primarily for short-term prediction. To extend CFT refinement toward EoL, a dynamic LNN is introduced to model the long-term degradation evolution. The dynamic LNN adopts a similar model configuration to the static one, but is trained on the CFT data throughout batteries' lifespans to learn the overall temporal dynamics. Instead of being adapted online like the static LNN, it refines degradation trajectories through a sliding-window process, e.g., over 100-cycle intervals, each of which captures a local segment of the long-term capacity evolution. This design allows the dynamic LNN to recognize degradation inflection points across different batteries and adaptively extend the CFT beyond early degradation stage. Overall, two LNNs operate in sequence in deployment. The static one first refines the CFT in early-stage using early-cycle information and entropy-guided features. When the SoH drops further, the dynamic LNN continues the refinement iteratively until EoL. Together, the design enables continuous, real-time, and physically consistent CFT refinement across different batteries and operating conditions.

\begin{figure}[]
\centerline{\includegraphics[width=\columnwidth]{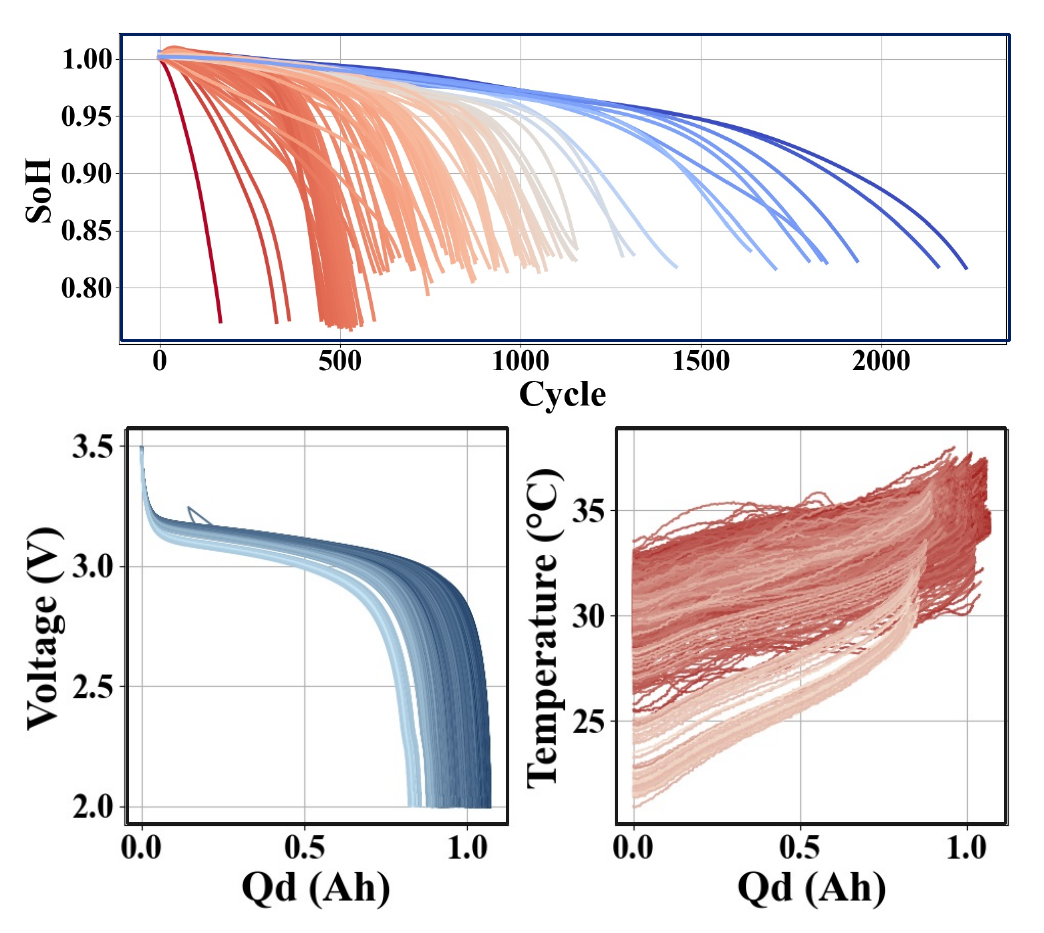}}
\caption{The original data extraction from the dataset, which consists of 124 batteries. The extracted data include the SoH, voltage, and surface temperature profiles.
}
\label{fig:data}
\end{figure}

\section{Dataset Preparation}
\label{sec:data}
The proposed \textsc{EntroLnn} is evaluated using the MIT-Stanford battery degradation dataset \cite{MIT_2019_dataset}, which comprises 124 commercial 18650-type lithium iron phosphate (LFP) batteries tested under various charging/discharging protocols. The batteries exhibit cycle lives ranging from few than 500 to more than 2,000 cycles, as shown in Figure \ref{fig:data}. We can also observe from the figure that voltage data has high quality and consistency, whereas the temperature measurements contain much noises and fluctuation and the initial temperature at the beginning of discharging is non-uniform. This is because of the reasons such as sensor bias and insufficient cooling, and we perform necessary data preprocessing such as normalization. We use \texttt{Bat003} as the reference battery for training static LNN. Among the batteries, \texttt{Bat003} has the longest lifespan with 2,234 cycles. A subset of batteries is used to train the dynamic LNN and the remaining batteries are reserved for operando adaptation and validation. We also adopt the thermal model of LFP batteries developed in existing studies for extracting entropy features.

\section{Experimental Study}
\label{sec:experiment}
In this section, the performance of \textsc{EntroLNN} is validated in two aspects, including pointwise SoH estimation based on operando entropy feature extraction, and long-term CFT prediction. A comparative study is further conducted to demonstrate the necessity and effectiveness of the proposed \textsc{EntroLnn}. All model training and evaluations are performed on a workstation equipped with an NVIDIA RTX 3500 GPU using CUDA 12.7.

\subsection{Operando Entropy-Based SoH Estimation}
\begin{figure}[]
\centerline{\includegraphics[width=\columnwidth]{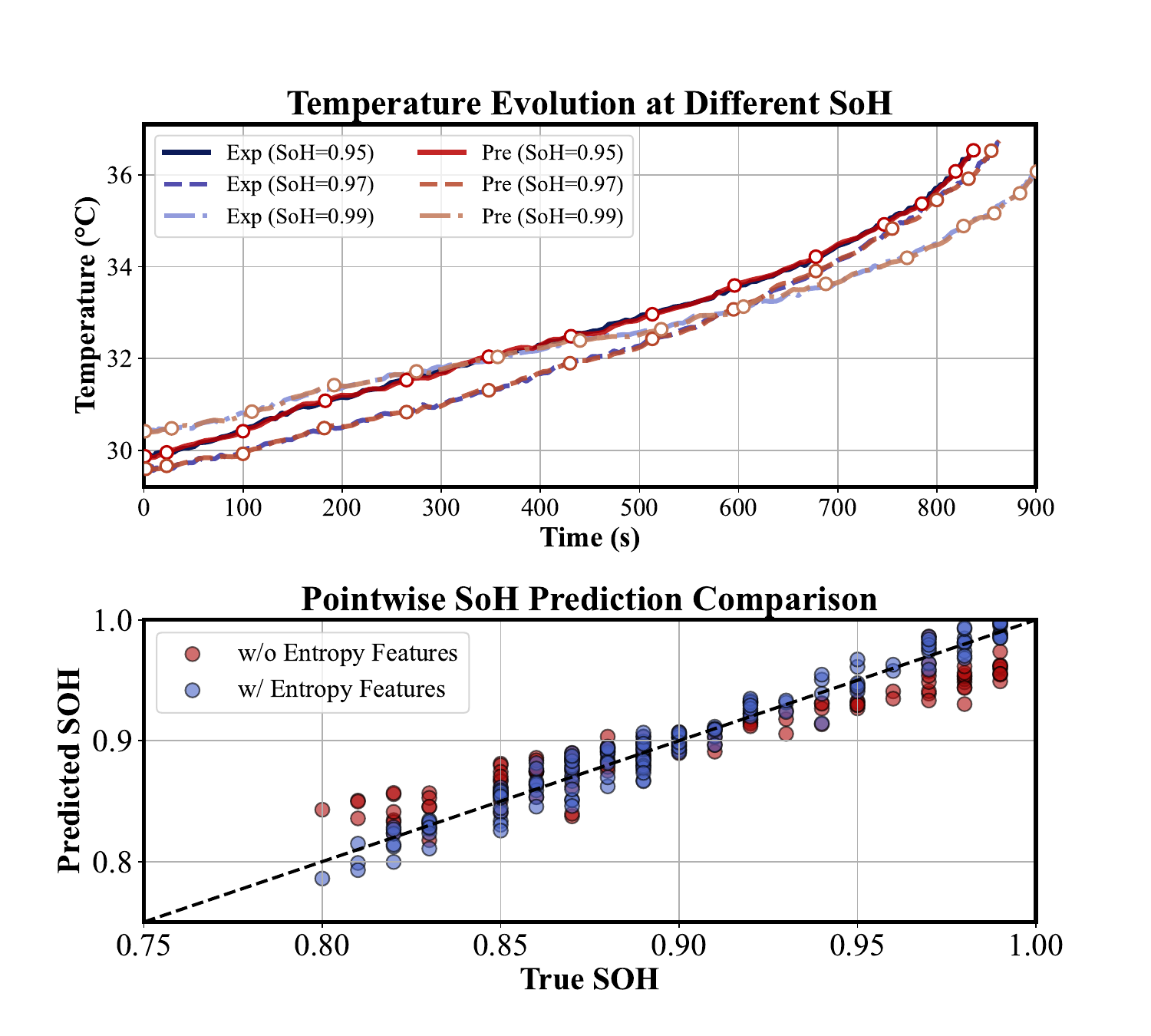}}
\caption{The operando temperature curve refinement and pointwise SoH estimation results. \textsc{EntroLnn} adapts to each battery and identifies the corresponding heat generation model. Without entropy feature guidance, the pointwise SoH estimation produces significantly higher errors. 
}
\label{fig:Pointwise_SOH}
\end{figure}

The entropy features based on temperature cannot be directly verified against raw sensor data. Due to the compact geometry and radial symmetry of cylindrical batteries, the internal temperature distribution typically exhibits a smooth and stable spatial pattern. Therefore, a good alignment between the calculated and measured surface temperatures is considered sufficient to indicate the fidelity of our modeled temperature field. Figure \ref{fig:Pointwise_SOH} shows the operando surface temperature refinement results, and our thermal model achieves an MAE of 0.0349 $^{\circ}$C and a mean absolute percentage error (MAPE) of 0.1\% across the tested SoH range. The results demonstrate that the thermal model and our online temperature curve adaptation method possess sufficient capability to enable accurate real-time entropy extraction. Notably, the MIT-Stanford dataset lacks a complete thermal model, and the initial heat generation model employed here was developed based on a 1.8 Ah 18650 LFP battery in a healthy condition. Despite this mismatch, the detailed results in Figure \ref{fig:Pointwise_SOH} reveal that the proposed online temperature adaptation framework is capable of compensating for discrepancies between the reference model and the actual battery behaviors. Specifically, it maintains temperature prediction errors below 0.0346 $^{\circ}$C, 0.0315 $^{\circ}$C, and 0.0386 $^{\circ}$C at SoH levels of 99\%, 97\%, and 95\%, respectively.

By combining temperature-based entropy with additional physics-informed indicators, \textsc{EntroLnn} achieves high-fidelity pointwise SoH estimation, as shown in Figure \ref{fig:Pointwise_SOH}. In this figure, the blue and red points represent the SoH estimation results with and without entropy features, respectively. For \textsc{EntroLnn} with entropy features, the MAE is only 0.0086, less than 1\% of the practical SoH. This indicates that the estimation results have sufficient fidelity for serving as a reference for the following short-term CFT refinement. Based on the comparison between two SoH estimation pathways, the temperature-based entropy further demonstrates its necessity on the pointwise SoH estimation, since the single temperature feature can only provide 0.0158 MAE, 1.8x higher than its counterpart. Moreover, all features are derived from temperature and voltage signals, and current measurements are not necessary. Thereby, \textsc{EntroLnn} has reduced sensor dependency and impact of sensor measurement noises. These results show that entropy features not only enhance SoH estimation accuracy but also improve sensor efficiency in operando battery analytics.

\subsection{Operando CFT Refinement Accuracy}
In most battery analytics studies, model performance is primarily evaluated through the accuracy of pointwise SoH estimation at individual cycles. These metrics overlook the continuous and dynamic nature of battery degradation. In practice, degradation evolves as a trajectory rather than isolated points. To capture this complete behavior, \textsc{EntroLnn} in this paper is evaluated based on CFT, which reflects the full temporal progression of SoH and degradation trends. A predicted CFT well-aligned with ground-truth shows that a model not only provides accurate SoH estimations but also reproduces the long-term degradation dynamics until EoL.

\begin{figure*}[!t]
\centering
\includegraphics[width=0.98\textwidth]{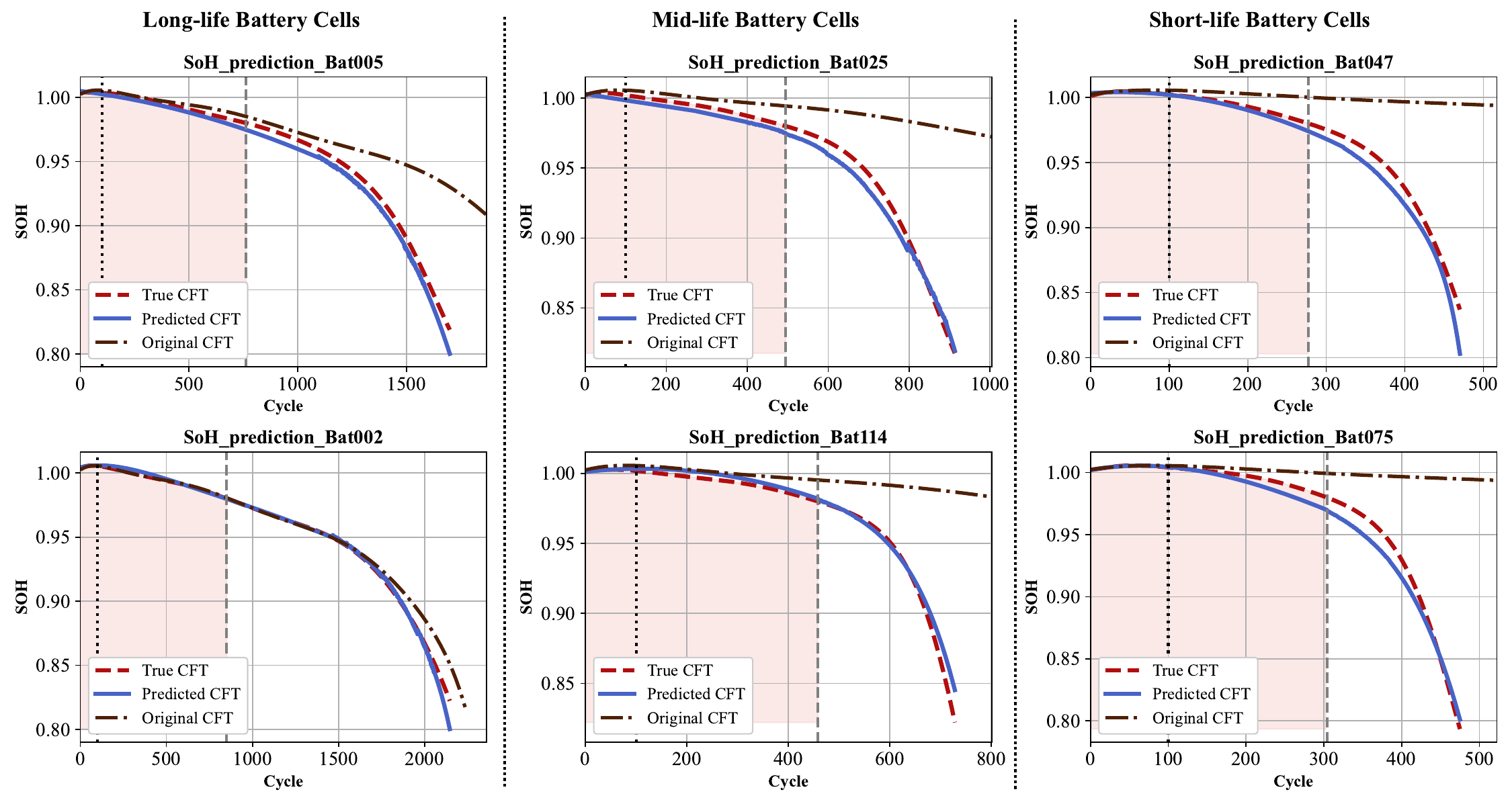}
\caption{CFT operando refinement for different batteries. The experimental results can be grouped into three categories corresponding to long-, mid-, and short-life batteries, to illustrate the transformable CFT refinement from the original dataset. The shaded region represents the SoH range from 100\% to 98\%. \textsc{EntroLnn} accurately models the CFTs across different batteries.
}
\label{fig:CFT refinement}
\end{figure*}
\begin{table*}[] 
\centering
\caption{Performance Comparison of Different Methods. MAE of SoH estimation and EoL prediction reveal the holistic accuracy of the \textsc{EntroLnn}. The data storage and model parameter number demonstrate the efficiency and the practical feasibility. }
\def\arraystretch{1.2}
\begin{tabular}{@{}ccccc@{}}
\toprule\toprule
\textbf{Method} & \textbf{MAE (SoH)} & \textbf{MAE (EoL)} & \textbf{Data Storage} & \textbf{Parameters (M)} \\
\midrule
BFRN  & 0.004775 & 200 (with 10\% data) & 0.36 GB for 10 cycles & 6.05 \\ 
BTL   & 0.02709 (with 30\% data) & -- & -- & -- \\ 
DRRN  & 0.012 & -- & -- & 0.69 \\ 
TLPH  & -- & 47.67 & -- & 7.36 \\ 
DCNN  & -- & 65.00 & -- & 2.39 \\ 
\textbf{\textsc{EntroLnn}} (ours) & \textbf{0.004577} (with 10\% data) & \textbf{18} (with 10\% data) & \textbf{239KB for 2,134 cycles} & \textbf{0.25} \\
\bottomrule\bottomrule
\end{tabular}
\label{tab:comparison}
\end{table*}

\subsubsection{CFT Refinement and SoH Evolution}
\textsc{EntroLnn} enables a two-stage refinement of the SoH evolution process, comprising short-term and long-term refinement. The former assesses the accuracy of the static LNN and its physics-informed performance, while the integration of dynamic LNN subsequently yields a robust long-term CFT refinement. In both stages, the emphasis is placed on achieving high fidelity with limited data and enhancing the model’s generalizability across different batteries. The physics-informed refinement process of \textsc{EntroLnn} facilitates seamless transformation from a reference battery to different batteries, as demonstrated in the beginning CFT refinement stage. As shown in the shaded region of Figure \ref{fig:CFT refinement}, which ends at 98\% SoH, the refined SoH evolution is entirely derived from the static LNN model trained solely on battery \texttt{Bat003}. Remarkably, the average MAE of the SoH evolution in the beginning stage across all 124 batteries reaches as low as 0.0025, relying only on the first 100 cycles and the enhanced pointwise SoH. It is worth noting that the ground-truth used for the static LNN spans 2,234 cycles. Even when applied to practical batteries with significantly shorter lifespans, such as \texttt{Bat047} with 470 cycles and \texttt{Bat075} with 475 cycles in Figure \ref{fig:CFT refinement}, \textsc{EntroLNN} remains capable of capturing the sharp degradation profiles in the early stage. The MAEs for these two batteries are 0.0018 and 0.0033, respectively.

Furthermore, after the static LNN provides accurate short-term SoH refinement up to 98\% of SoH, the dynamic LNN is integrated to capture inflection points and deliver accurate long-term predictions. Similar to the short-term refinement, the MAE of the entire CFT remains as low as 0.0045. As illustrated in Figure \ref{fig:data}, the degradation trajectories of short-lifespan batteries diverge significantly from those of long-lifespan counterparts. Nevertheless, the combined static-dynamic LNN framework maintains an MAE below 0.0058 for batteries with lifespans under 800 cycles, with the corresponding MAPE reduced to only 0.6\%. Comparison with the results in the foregoing section indicates that the long-term MAE is slightly higher than that of the short-term stage, primarily due to the diminishing influence of physics-informed entropy features beyond an SoH of 98\%. The inflection point of the CFT is a critical feature, but in our framework it is mainly inferred from the Markovian dynamics of SoH evolution between sequential 100-cycle segments. Based on our analysis, even with only online physics data available before the SoH reaches 98\%, the model can already predict over the remaining CFT accurately.

\subsubsection{EoL Prediction Accuracy}
The general architecture for EoL prediction typically achieves high accuracy with sufficient historical data. However, in practical applications, accurately estimating EoL at an early stage with limited data is crucial for effective battery management. In \textsc{EntroLnn}, EoL prediction is performed using only the first 100 cycles and additional data up to an SoH of 98\%, which corresponds to approximately 10\% of a battery's full lifespan. Based on the results from 123 batteries (excluding the reference battery \texttt{Bat003}), the average MAE of EoL prediction is only 18 cycles, with a corresponding MAPE of 1.49\%. Several representative cases are shown in Figure \ref{fig:CFT refinement}, demonstrating that the maximum prediction uncertainty of \textsc{EntroLnn} is only 2.17\%. Moreover, early-stage EoL prediction is inherently challenging, particularly when the target battery differs significantly from the training reference. Following previous studies, the MIT-Stanford dataset can be categorized into three groups for long-, mid-, and short-life batteries. As shown in Figure \ref{fig:CFT refinement}, \texttt{Bat002} and \texttt{Bat005} are long-life examples; \texttt{Bat025} and \texttt{Bat114} represent mid-life cases; \texttt{Bat047} and \texttt{Bat075} are short-life batteries. Using \texttt{Bat003} as the sole reference, the MAEs for EoL prediction of the two mid-life batteries are 3 and 14 cycles, respectively. Even for the more distinct short-life batteries, the MAEs remain small, i.e., 10 and 3 cycles. These results confirm that \textsc{EntroLnn} is capable of transferable and reliable EoL prediction, even with limited and not fully representative data.

\subsection{Comparison Study}
In the context of operando refinement of the CFT, it is imperative to ensure both high predictive fidelity and robust generalizability across heterogeneous battery systems. Moreover, the computational efficiency of different ML architectures is still important, particularly for real-time and embedded battery applications. As summarized in Table \ref{tab:comparison}, a range of advanced ML models have been developed for CFT refinement \cite{Nian_Cai, Jiarui_Zhang}. Among them, the Bimodal Fusion Regression Network (BFRN) \cite{Jiarui_Zhang} stands out as the only framework that delivers comprehensive CFT refinement for both SoH trajectory reconstruction and EoL prediction. For standalone SoH estimation tasks, the Bayesian Transfer Learning (BTL) framework \cite{Nian_Cai} and the Dilated Residual Regression Network (DRRN) achieve promising accuracy, with 1\%–3\% MAPE using about 30\% of the data from each single battery. Besides, models such as the Transfer Learning Parallel Hybrid (TLPH) and the Dilated Convolutional Neural Network (DCNN) report only EoL estimation performance, which is less competitive compared to our \textsc{EntroLnn}. Although BFRN yields comparable results in terms of CFT refinement, it lacks integration with physics-informed entropy, thereby leading to low EoL prediction accuracy, particularly under data-constrained scenarios, e.g., with only 10\% data. In contrast, \textsc{EntroLnn} maintains high EoL prediction fidelity even under such data scarcity conditions, demonstrating superior operando adaptability with entropy. Furthermore, BFRN's reliance on high-dimensional image-based modules introduces substantial storage overheads, e.g., approximately 0.36 GB to store just 10 cycles of battery data. By comparison, \textsc{EntroLnn} accommodates complete 2,134-cycle CFT refinement with only 239 KB of data storage, which underscores its practicality for resource-constrained and real-time deployment. The remarkable computational efficiency of \textsc{EntroLnn} is primarily attributed to its usage of physics-based entropy and adoption of LNNs, which are inherently suited for modeling temporal dependencies and preserving memory of dynamic processes. As a result, all baseline models listed in Table \ref{tab:comparison} exhibit significantly higher numbers of model parameters. For example, BFRN requires nearly 23$\times$ more parameters to approximate \textsc{EntroLnn}'s fidelity, highlighting \textsc{EntroLnn}'s efficiency and scalability.

\section{Conclusion}
\label{sec:conclusion}
This study extends the traditional pointwise SoH estimation to full CFT refinement. We introduce \textsc{EntroLnn}, a framework with entropy-guided features and transformable LNNs designed to achieve high generalizability and fidelity with minimal computational overhead. Using a publicly available battery dataset for validation, the results show that temperature-based entropy serves as a meaningful operando feature for modelling the entire CFT. In the proposed \textsc{EntroLnn}, the real-time thermal model transfers adaptively across different batteries, achieving a temperature calculation MAE of 0.0349 $^{\circ}$C only. The high fidelity entropy features extracted subsequently enable precise SoH estimation, with an average MAPE of 0.1\%. Comprehensive evaluation of the entire CFT process shows that \textsc{EntroLnn} achieves MAEs of 0.004577 for CFT refinement and 18 cycles for EoL prediction, respectively, for different batteries. The comparative analysis further shows that \textsc{EntroLnn} delivers competitive predictive accuracy with a lightweight architecture that requires minimal data and computation than conventional methods. Overall, this study provides a practical pathway toward self-adaptive, interpretable, and deployable battery health analytics for next generation battery management systems.


\begin{acks}
This research is supported by A*STAR under its MTC Programmatic (Award M23L9b0052), MTC Individual Research Grants (IRG) (Award M23M6c0113), SIT’s Ignition Grant (STEM) (Grant ID: IG (S) 2/2023 – 792), and the National Research Foundation Singapore and DSO National Laboratories under the AI Singapore Programme (AISG Award No: AISG2-GC-2023-006).
\end{acks}

\printbibliography

\appendix
\end{document}